\documentclass[twocolumn,final]{elsarticle}
\usepackage[colorlinks=true,linkcolor=blue]{hyperref}
\usepackage{lineno}
\usepackage{multirow}
\usepackage{hyperref}
\usepackage{soul}
\usepackage[normalem]{ulem}
\usepackage[dvipsnames]{xcolor}
\modulolinenumbers[1]
\journal{Computers in Biology and Medicine}

\biboptions{sort&compress}
\begin{document}
\begin{frontmatter}
\title{COVID-19 forecasting using new viral variants and vaccination effectiveness models}
\author[a]{Essam A. Rashed}
\ead{rashed@gsis.u-hyogo.ac.jp}
\author[b]{Sachiko Kodera}
\author[b,c]{Akimasa Hirata}
\address[a]{Graduate School of Information Science, University of Hyogo, Kobe 650-0047, Japan}
\address[b]{Department of Electrical and Mechanical Engineering, Nagoya Institute of Technology, Nagoya 466-8555, Japan}
\address[c]{Center of Biomedical Physics and Information Technology, Nagoya Institute of Technology, Nagoya 466-8555, Japan}


\begin{abstract}

Recently, a high number of daily positive COVID-19 cases have been reported in regions with relatively high vaccination rates; hence, booster vaccination has become necessary. In addition, infections caused by the different variants and correlated factors have not been discussed in depth. With large variabilities and different co-factors, it is difficult to use conventional mathematical models to forecast the incidence of COVID-19. Machine learning based on long short-term memory was applied to forecasting the time series of new daily positive cases (DPC), serious cases, hospitalized cases, and deaths. Data acquired from regions with high rates of vaccination, such as Israel, were blended with the current data of other regions in Japan such that the effect of vaccination was considered in efficient manner. The protection provided by symptomatic infection was also considered in terms of the population effectiveness of vaccination as well as the vaccination protection waning effect and ratio and infectivity of different viral variants. To represent changes in public behaviour, public mobility and interactions through social media were also included in the analysis. Comparing the observed and estimated new DPC in Tel Aviv, Israel, the parameters characterizing vaccination effectiveness and the waning protection from infection were well estimated; the vaccination effectiveness of the second dose after 5 months and the third dose after two weeks from infection by the delta variant were 0.24 and 0.95, respectively. Using the extracted parameters regarding vaccination effectiveness, DPC in three major prefectures of Japan were replicated. The key factor influencing the prevention of COVID-19 transmission is the vaccination effectiveness at the population level, which considers the waning protection from vaccination rather than the percentage of fully vaccinated people. The threshold of the efficiency at the population level was estimated as 0.3 in Tel Aviv and 0.4 in Tokyo, Osaka, and Aichi. Moreover, a weighting scheme associated with infectivity results in more accurate forecasting by the infectivity model of viral variants. Results indicate that vaccination effectiveness and infectivity of viral variants are important factors in future forecasting of DPC. Moreover, this study demonstrate a feasible way to project the effect of vaccination using data obtained from other country.

\end{abstract}


\end{frontmatter}



\section{Introduction}

The emergence of Coronavirus Disease-2019 (COVID-19) in late 2019 resulted in several changes in the daily routine of people and has become a significant cause of mortality worldwide, causing more than 5.9 million deaths~\cite{who2022}. Due to vaccination, the number of daily positive cases (DPC) has decreased in several countries. Although some countries have achieved high vaccination rates~\cite{Lurie2020NEJM}, other countries are far behind, with only a small proportion of their respective populations being vaccinated. This is mainly due to the lack of resources~\cite{Wouters2021challenges}, vaccination hesitancy~\cite{machingaidze2021understanding,ALAMOODI2021104957}, or other related issues.

One of the first countries to vaccinate its population was Israel; however, relatively high DPC were reported in August 2021 despite the country’s vaccination rate being above 68\%~\cite{mizrahi2021correlation}. One reason for this upsurge was attributed to the high transmissibility of the Delta variant~\cite{wadman} and the waning protection from vaccination~\cite{khoury2021neutralizing}, especially for those who have been vaccinated very early during the pandemic~\cite{sanderson2021covid}. A similar trend was observed in the United Kingdom~\cite{pouwels2021effect} and the United States~\cite{cohn2021breakthrough}. The data obtained from countries with high vaccination rates would be useful in predicting the future potential in follow-up regions. However, it is difficult to manipulate data acquired from other regions considering the variations of the different influencing factors.

Here, we propose an efficient method based on deep learning framework to forecast COVID-19 of one area based on viral variant modeling and vaccination effectiveness using a framework for data projection obtained from other regions. Considering that more than 65\% of countries still have vaccination rates below 70\% [12], the proposed approach can learn the experience of countries with high vaccination rates in a way that can provide useful insights for other countries with low vaccination rates. Moreover, the developed model can provide a clearer understanding of potential booster shot requirements and when they should be administered. The contribution of this work can be summerized as follows:
\begin{itemize}
\item Development of new model that enable projection of vaccination effectiveness at population level from one country to another.
\item Construct viral infectivity model that enable different pandemic spread considering percentage of the variant and potential relative infectivity.
\item Optimize model parameters using data from Tel Aviv, where vaccination rate is relative high compared with other countries.
\item Validation study of different data inputs in the accuracy of DPC forecasting within different pandemic waves such as spread, peak, and decay phases for three prefectures in Japan.
\end{itemize}

The remainder of this article is structured as follows. Section 2 discuss related work with emphasis on machine learning/deep learning approaches. In Section 3, the proposed method is discussed while data description is presented in Section 4. Different scenarios are discussed in Section 5 and achieved results are Section 6. Discussion and conclusion are in Sections 7 and 8, respectively.

\section{Related work}

Several national and regional projects are currently in progress to predict the infection during the COVID-19 pandemic~\cite{covidaijp,ihme2020modeling}. The aim of such projects is to understand the pandemic data to improve medical resource allocation, intervention, and policy settings. Susceptible, Exposed, Infectious, and Recovered (SIR or SEIR) models have often been used to solve these problems~\cite{he2020seir,carcione2020simulation}, and several recent studies have demonstrated the robust abilities of machine learning approaches to adjust for realistic scenarios without forming strict assumptions~\cite{ZEROUAL2020110121}. In contrast, earlier studies did not consider the public’s mobility~\cite{a13100249}, which has been clarified as a dominant factor characterizing new cases as a surrogate marker for social distancing~\cite{chang2021mobility,ijerph18115736}. In addition, the forecasts were limited to only a few days~\cite{nikparvar2021spatio}. Recently, several studies have considered the data pattern change due viral variants, and we have underlined the difficulty of predicting when new variants will appear~\cite{ijerph18157799}. It is critical for any successful model to be able to predict the beginning of a new wave of infections and its potential magnitude. The difficulty in modeling the upsurge of cases may also be attributed to behavioral changes when the DPC are low. Machine learning and deep learning models was used for knowledge discovery and forecasting of different aspects associated with the COVID-19 pandemic~\cite{XU2022105342,RASHED2021103743,9099302,CHIMMULA2020109864,Gatta9234698}. A systematic review that summerize recent work on COVID-19 data analysis is in~\cite{HEIDARI2022105141}. 

Although conventional projection strategies did not thoroughly consider the effects of vaccination, some studies did~\cite{WINTACHAI2021e06812}. With the emergence of messenger RNA vaccines, the efficiency of vaccination in protecting against infection will dramatically improve. The weekly incidence of COVID-19 since administering the first vaccine dose started to decrease after two weeks, which further decreased after four weeks~\cite{doi:10.1056/NEJMc2101951}. After the second dose, vaccine effectiveness reached 75\%–95\% after a few weeks~\cite{zeng2021effectiveness,doi:10.1056/NEJMc2114290}. However, vaccine efficiency may depend on the dominant viral variant~\cite{zeng2021effectiveness}. For the Pfizer-BioNtech vaccine, its efficacy for health-care workers has been extensively examined~{\cite{doi:10.1056/NEJMc2101951}} and complete vaccination is defined as two doses given 21 days apart. Therefore, data from regions with high vaccination rates and new variants may serve as important guides for forecasting potential risks in other regions.

\section{Methods}
\subsection{Forecasting model}

\begin{figure*}
\centering
\includegraphics[width=.9\textwidth]{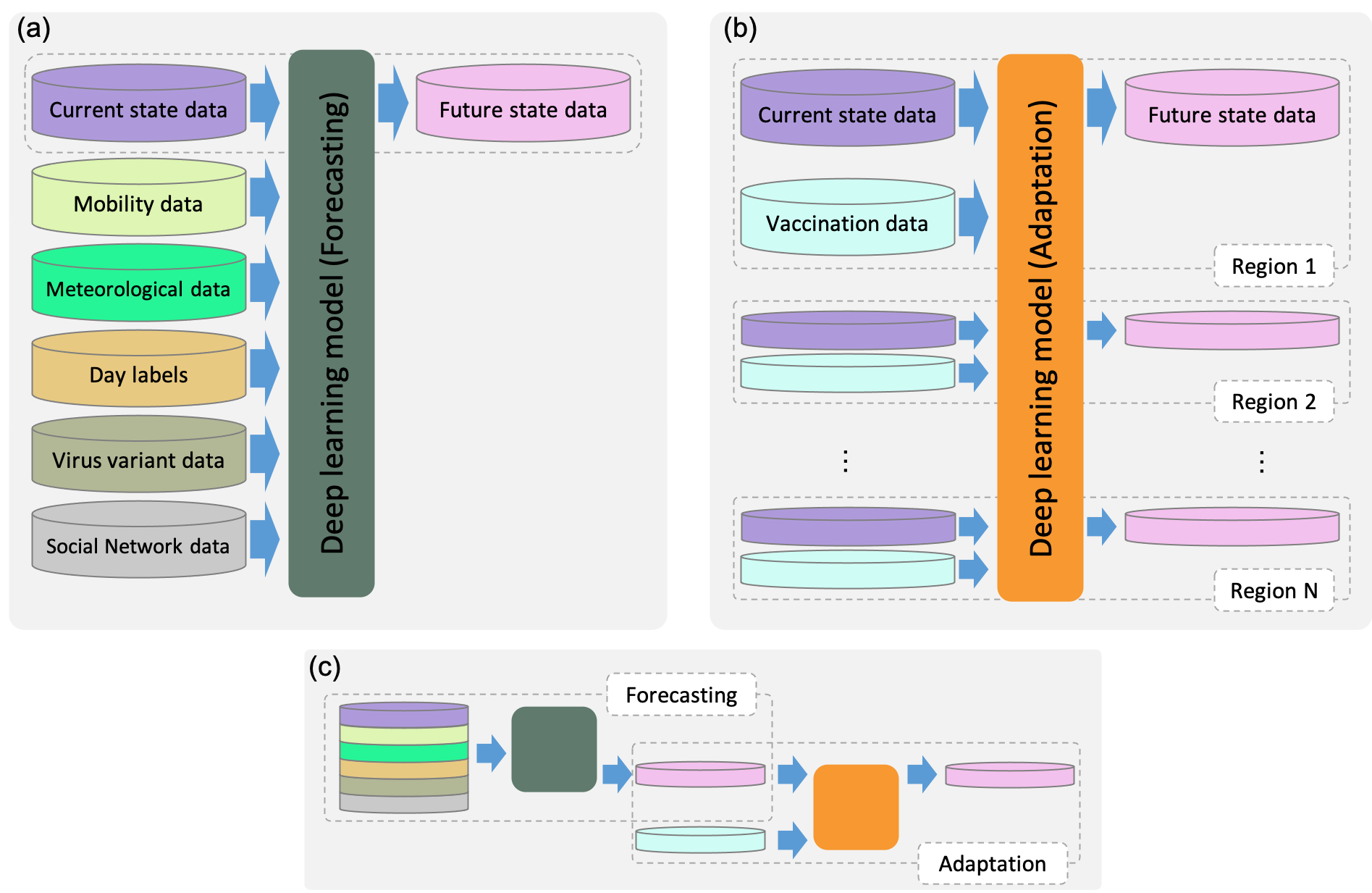}
\caption{Outline of proposed model for COVID-19 forecasting with vaccination effectiveness. (a) Initial forecasting is computed using a blend of time-series  data; (b) the vaccination effect is computed using data acquired from different regions; and (c) the full model includes steps in (a) forecasting and (b) adaptation. Network detailed architecture is in Fig.~\ref{fig2}.}
\label{fig1}
\end{figure*}

\begin{figure*}
\centering
\includegraphics[width=0.9\textwidth]{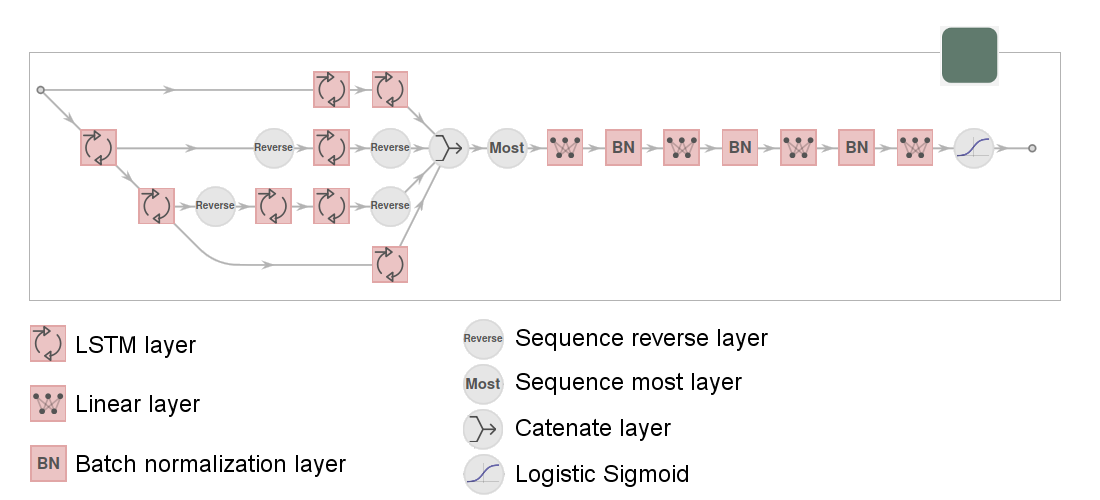}
\caption{LSTM network architecture.}
\label{fig2}
\end{figure*}

\begin{figure*}
\centering
\includegraphics[width=0.9\textwidth]{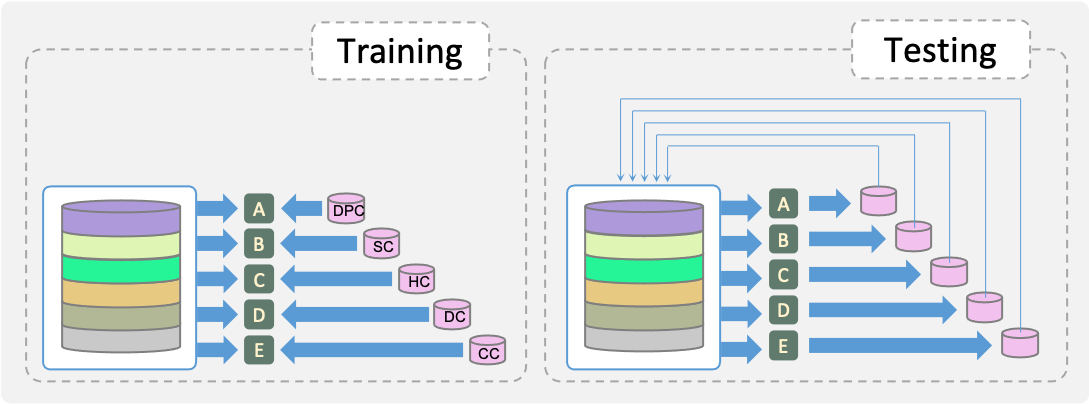}
\caption{Training and testing phases for the COVID-19 forecasting model. In training, different networks (A–E) are trained to forecast specific indicators. Long-term forecasting is achieved in the testing phase with concurrent data updates.}
\label{fig3}
\end{figure*}

The forecasting model was designed using a multi-path long short-term memory (LSTM) neural network based on our earlier study detailed in~\cite{ijerph18157799}. The main difference of LSTM from conventional methods such as SIR/SEIR is that in the deep learning model, the number of variables (network features) are extremely large and can handle data non-linearity in a more efficient manner. The network parameters are optimized based on an ablation study~\cite{ijerph18157799}. The main data stream of the forecasting model and fine details on the training and testing phases are shown in Fig.~\ref{fig1} and network detailed architecture is in Fig.~\ref{fig2}. This architecture is implemented in Wolfram Mathematica (R) version 12.1 installed on a workstation with four Intel (R) Xeon CPUs running at 3.60 GHz, with 128 GB of memory and three NVIDIA GeForce 1080 GPUs. Training is implemented through a set of networks, with each network trained to forecast a single indicator (DPC, serious cases [SC], hospitalized cases [HC], daily death cases [DC], or daily hospital discharged cases [CC]). The current models are trained to forecast different COVID-19 incidences (DPC, SC, HC, DC and CC) for 14 future days. Then, the estimated values are used as input again for further 14 days (day 15 to day 28) forecasting (recurrent data), and so on. This feature is demonstrated in Fig.~\ref{fig3} (Testing) where big arrows indicate the normal flow of data testing to get the estimation of (day 1 to day 14). The estimated values are feedback (small arrows) as input for forecasting of (day 15 to day 28) and further future estimation.

\subsection{Adaptation model}

The projection of the epidemic tendency in one country to other countries is not always successful as epidemic parameters and associated factors in different countries may not be consistent. The two models work well especially during the early stages of vaccination when the effects of vaccination are still unclear. The adaptation model is trained using a simplified combination of data wherein the target must learn the effects of vaccination within different stages of the pandemic. With this design, we can think of the \emph{forecasting model} as the local scope network and the \emph{adaptation model} as the global scope network. This strategy can efficiently enable the use of data of countries with high vaccination rates without considering local features. The data flow between the forecasting and adaptation models is shown in Fig.~\ref{fig1} (c).

\subsection{Vaccination effectiveness at population level}

As the vaccination rate may not reflect the actual efficiency due to the variations in vaccines and waning protection over time~\cite{sanderson2021covid}, we proposed a representation of vaccination protection, which is defined as the vaccination effectiveness at the population level that considers the waning protection and is used as a metric for herd immunity. The vaccination effectiveness at the population level in each city or prefecture was assumed to be as follows:

\begin{equation}
E(d)=\sum_{i=0}^{d} \frac{\sum_{t=1}^{T} (N_t(d-i)e_t(i))}{ P},
\label{eq1}
\end{equation}

\begin{equation}
e_t(i)=\max(0,\tilde{e}_t(i)),~~\tilde{e}_t(i)=\left\{\begin{array}{ll}
a_t \frac{i}{K} & (i \leq K)\\
a_t-s(i-K) & (i>K)\\
\end{array}\right.
\end{equation}
where $d$ is the day index and $N_t(d-i)$ denotes the number of people newly administrated with the $t$\textsuperscript{th} vaccine dose on day ($d-i$). $e_t(i)$ is the non-negative individual vaccination effectiveness of the $t$\textsuperscript{th} dose on $i$ days after inoculation. Parameters $a$ and $s$ are adjusted to reach an individual vaccination effectiveness peak within $K$ days after inoculation, which then decrease linearly due to waning effect~\cite{tartof2021,andrews2021}. The wanning effect was adjusted when people inoculated the second or third dose by considering the number of people vaccinated in the past (e.g., the number of second shot vaccinated people in past was reduced with increasing the number of people with third shot). To highlight this point an illustration demonstrate a simple example of population vaccination with different status of potential subjects is in Fig.~\ref{fig4}. $P$ is the population within urban region under consideration ($P$ is considered a constant value during the time frame of this study).

We assumed $T$ = 3 to demonstrate the number of vaccine doses (vaccines with a single dose, such as that of Johnson \& Johnson, was not considered here) and $K$ = 14 for the two-week latency period of the vaccination effect~\cite{Bernal2021NEJM}. The parameters of $a_1$ and $a_2$, characterizing the individual effectiveness of vaccination, for the Delta variant were chosen as 0.605 and 0.756, respectively, both of which are based on a meta-analysis of systematic review (11 study groups) as detailed in~\cite{zeng2021effectiveness}. These parameters coincide with the reported real-world vaccination effectiveness~\cite{aran2021estimating} and also consistent with computational estimation in Japan~\cite{Kodera2022vaccines}. The parameters of $s$ and $a_3$ are computed as explained below (please refer to Section 6.1). The antibody levels of infected people are comparable to or somewhat lower than those of fully vaccinated persons~\cite{doi:10.1056/NEJMoa2109072}; thus, the DPC is counted as additional value of fully vaccinated people.

\subsection{Infectivity of viral variant}
Different viral variants and mutations have been reported during the recent waves of infection of COVID-19. In addition, different variants have different rates of spread, infectivity, and resistance to the currently administered vaccines. This effect has become significant since the emergence of the Delta and Omicron variants. Therefore, developing an infectivity model based on the dominant (or partially spreading) variant is necessary. The normalized infectivity index ($\tilde{f}$) is computed using the following equation:

\begin{equation}
\tilde{f}(d)=(\beta-\alpha) \frac{f(d)-\min(f)}{\max(f) -\min(f)}+\alpha
\label{eq3}
\end{equation}

\begin{equation}
f(d)=\sum_{j=1}^{N}\omega_jv_{dj} ~~\forall d
\end{equation}
where $v_{dj}$ is the percentage of variant $j$ at day $d$, $\omega_j$ is a weighting parameter assigned to each variant that demonstrates its relative infectivity, and parameters $\alpha$ and $\beta$ are scaling parameters and $\max(f)$/$\min(f)$ are global values computed using all available measured data. As variant data were reported weekly, daily values were computed using linear interpolation. Therefore the normalized index of infectivity is an indicator of infectivity risk considering the percentage of different variant within study area.

\subsection{Validation measurements}

For quantitative evaluation, the average relative error over a period of $N$ days was computed as follows:

\begin{equation}
Error=\frac{1}{N}\sum_{d=1}^{N}\frac{|y_d-\hat{y}_d|}{y_d}
\end{equation}
where $y_d$ and $\hat{y}_d$ are the real and estimated positive cases on day $d$, respectively. 

\begin{figure*}
\centering
\includegraphics[width=0.9\textwidth]{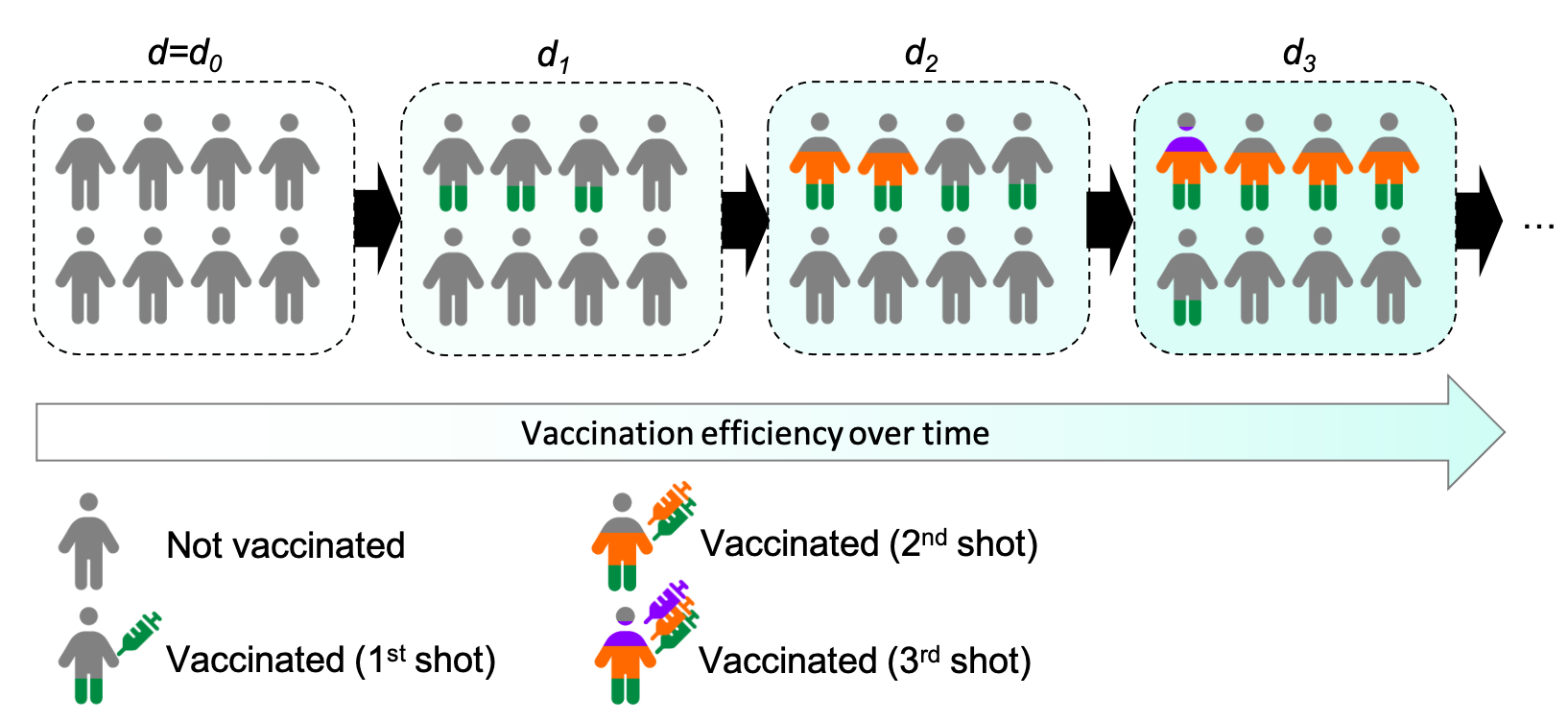}
\caption{Schematic explanation of the change in vaccination status with a sample population ($P$=8) over time. At $d=d_0$ $N_1$=$N_2$=$N_3$=0. At $d_1$, $N_1$=3 and $N_2$=$N_3$=0, at $d_2$, $N_1$=$N_2$=2 and $N_3$=0. Finally, at $d_3$, $N_1$=1, $N_2$=3 and $N_3$=1.}
\label{fig4}
\end{figure*}
\section{Data}

The data combination used in the forecasting model includes the (i) current COVID-19 parameters (DPC, SC, HC, DC and CC), (ii) mobility data (retail \& recreation, grocery \& pharmacy, parks, transit stations, workplaces, and residences), (iii) meteorological data (daily maximum and minimum temperatures and daily average humidity), (iv) day labels (working days or holidays [i.e., weekends or national holidays]), (v) viral variant infectivity, and (vi) vaccination effectiveness. The main difference of this model from that from our previous study is that serious cases, hospitalized cases, and deaths were considered in item (i) in addition to items (v) and (vi). In addition, in the analysis of Tokyo, Osaka, and Aichi, the number of tweets and population at night were considered, which are potentially related to changes in public behavior. The effectiveness of the latter can be reported in~\cite{nakanishi2021site}. The breakdown and definition of each dataset are listed in Table~\ref{tab1}. Vaccination data, along with current COVID-19 data, were collected from external regions (Tel Aviv, Israel).

\subsection{Input Data for Japan}

The COVID-19 data of Tokyo were obtained from online open sources maintained by the Japanese Ministry of Health, Labor, and Welfare (MHLW)\footnote{\href{https://www.mhlw.go.jp/stf/seisakunitsuite/bunya/0000164708\_00079.html}{MHLW open data}}. Mobility data were downloaded from the global Google mobility reports\footnote{\href{https://www.google.com/covid19/mobility/}{https://www.google.com/covid19/mobility/}}. Meteorological data were obtained from the Japan Meteorological Agency\footnote{\href{https://www.jma.go.jp/jma/indexe.html}{https://www.jma.go.jp/jma/indexe.html}}. Day labels were based on the Japanese calendar, which were assigned as 1/0 for working/vacation days, respectively. Official state-of-emergency declarations by the Japanese government were assigned as 1/0 for active/inactive status, respectively. Information about the dominant variant was obtained from official MHLW reports\footnote{\href{https://www.mhlw.go.jp/stf/seisakunitsuite/newpage\_00054.html}{MHLW Reports}}. Vaccination rates were obtained from the Government CIO’s Portal, Japan\footnote{\href{https://cio.go.jp/c19vaccine\_opendata}{https://cio.go.jp/c19vaccine\_opendata}}.

In several earlier studies, public mobility was used as a key indicator for public interaction and social distancing~(e.g.\cite{NOLAND2021101016,chang2021mobility,ijerph18115736}. However, mobility data was criticized as it may not clearly indicate the social behaviour, such as drinking parties, that is reported to be a potential major source of infection in Japan. Social networking data were obtained from Twitter, and mobility at downtown areas were computed as the nighttime population who stayed near restaurants and bars. Twitter data were used as it may indicate social activities where close contact occurs, and the downtown population was considered as several domestic reports have indicated that the main infection clusters may be due to close contacts in these areas. Tweets with keywords \emph{BBQ}, \emph{drinking party}, and \emph{karaoke} (in Japanese) were chosen as risk-related terms. Data were obtained from NTT Data, INC. and processed by Toyoda Lab., University of Tokyo and shared through the Cabinet Secretariat COVID-19 AI Simulation Project~\cite{covidaijp}. When determining the number of tweeted keywords, those completed during the day or the previous day, or those planned until the next day, were extracted. While it is difficult to confirm if these gathering events are actually hold or not, recorded data can clearly indicate time frames where these events are more popular. For corresponding tweets, information on the prefecture was extracted from the user’s address. Note that due to the limited number of tweets, we only focused on three prefectures (Tokyo, Osaka, and Aichi); the number of tweets in other prefectures was generally lower, and the required number of tweets from other prefectures was not obtained. This is one reason why the analysis focused on these prefectures only. Three (Tokyo), two (Osaka), and one (Aichi) metropolitan areas were selected to represent the downtown districts with restaurants and bars (mesh size of 500m$\times$500m area) (see, Fig.~\ref{fig4}).

\subsection{Input Data for Tel Aviv}

The COVID-19 data and vaccination rate in Tel Aviv were obtained from online open sources\footnote{\href{https://info.data.gov.il/datagov/home/}{https://info.data.gov.il/datagov/home/}}, and mobility data were obtained from Google mobility reports. The average interval between the administration of the first and second doses was assumed as three weeks.

\begin{figure*}
\centering
\includegraphics[width=0.5\textwidth]{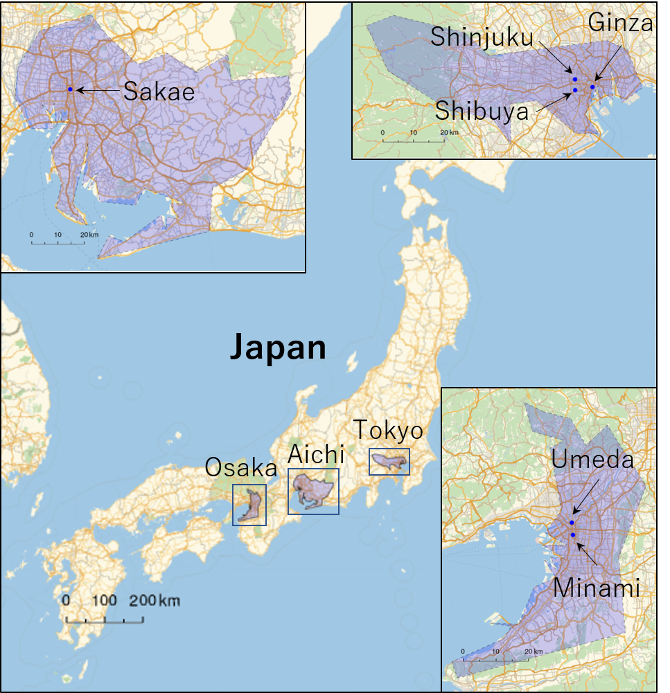}
\caption{Map of Japan with study areas and regions used to represent downtown districts.}
\label{fig5}
\end{figure*}

\section{Scenarios}
\subsection{Optimize vaccination effectiveness using Tel Aviv data} 
The vaccination effectiveness calculated from Eq. (\ref{eq1}) represents the model of protection from infection resulting from vaccination. With a variety of vaccines and other policy variables, parameters $a_t$ and s should be adjusted based on real local data. For this purpose, we replicated the DPC in Tel Aviv and adjusted the parameters. Tel Aviv was selected as it has a high vaccination rate and shared similarities with the vaccines used in Tokyo (BNT162b2). We then investigated three values for $s$ (0.21, 0.24, and 0.27), which characterizes the waning protection from vaccines, and the efficiency of the third dose (booster) was represented by $a_3$  (0.75, 0.85, and 0.95). A verification study using training data from August 1, 2020 to July 30, 2021 and testing data from August 1 to September 23, 2021 was conducted to estimate s and $a_3$. The optimum $a_3$ value was 0.95, which is consistent with that in the report of Pfizer and BioNTech~\footnote{\href{https://www.pfizer.com/news/press-release/press-release-detail/pfizer-and-biontech-announce-phase-3-trial-data-showing}{Pfizer and BioNTech Announce Phase 3 Trial Data[Accessed, March 5, 2022]}}. Also, the slope of the waning protection was 0.24, which agrees with large scale study~\cite{cohn2021breakthrough}.

\begin{table*}
\centering
\footnotesize
\caption{Datasets used in the forecasting/adaptation models shown in Fig.~\ref{fig1}.}
\setlength{\tabcolsep}{3pt}
\begin{tabular}{ c l l l }
\hline
\#&Dataset&Items &Scale\\
\hline
1&Current state data &1-1 Positive cases  & Daily number of cases  \\
&&1-2 Serious cases&\\
&&1-3 Hospitalized cases&\\
&&1-4 Deaths&\\
&&1-5 Hospital discharged cases&\\
\hline
2&Mobility data &2-1 Retail \& recreation  & Percent change from baseline (pre-pandmic)  \\
&&2-2 Grocery \& pharmacy&\\
&&2-3 Parks&\\
&&2-4 Transit stations&\\
&&2-5 Work places&\\
&&2-6 Residents&\\
&&2-7 Downtown area population&Daily number of persons\\
\hline
3&Meteorological data &3-1 Maximum temperature & Daily value \\
&&3-2 Minimum temperature&\\
&&3-3 Average humidity&\\
\hline
4&Day labels & 4-1 Working/holiday/ext. holiday & Labels (0/1/2) \\
&&4-2 Normal/State of emergency&\\
\hline
5&Variant infectivity & 5-1 $\tilde{f}(d)$ & Computed using Eq.(\ref{eq3}) \\
\hline
6&Behaviour &6-1 Tweets (\emph{nomikai}) & Daily tweets using keywords  \\
&&6-2 Tweets (\emph{karaoke}) &\\
&&6-3 Tweets (\emph{BBQ})&\\
&&6-4 Downtown area population&Daily number of persons\\
\hline
7&Vaccination effectiveness & 7-1 $E(d)$ & Computed using Eq.(\ref{eq1})\\
\hline
\end{tabular}
\label{tab1}
\end{table*}

\subsection{Exploring different input parameters for Tokyo, Osaka, and Aichi}
We then explored parameters which will correlate well with the new DPC. The main factors that might potentially influence the DPC in the future are listed in Table~\ref{tab1}. The viral variant is essential and can be extracted from the data in different countries, and the day label of “holiday” is potentially related to behavioral changes; both of which can be easily defined with any uncertainty, thus their use as the default parameters. There are different categories for mobility, including those in different urban regions. In our previous study, we have shown that in most prefectures, mobility at transit stations is an essential parameter, whereas remaining is also related to public activities in different urban regions. In addition, the nighttime population can identify social activities in regions where infection clusters were reported, which we compared using a new set of input parameters. Although weather was suggested to correlate with the number of DPC in some studies, others have denied this~\cite{majumder2021systematic}. In this study, temperature and humidity, which are also related to the absolute humidity, were considered simultaneously. The vaccination effect was considered in the adaption of the neural network. To demonstrate the effectiveness of our proposed forecasting system, we applied the same scenarios for Tokyo to Osaka and Aichi. Input parameter optimization was then conducted to validate the accuracy of forecasts. 

\section{Results}

\subsection{Extraction of parameters characterizing the waning protection from vaccination and third dose}

The parameters in Eq.~(\ref{eq1}) were revised to replicate new DPC in Tel Aviv. From Fig.~\ref{fig5}, the observed and replicated DPC were in agreement when $s$ and $a_3$  were set to 0.27 and 0.95, respectively. The average residual error of DPC from August 27 to November 21, 2021 was 0.289. Considering the incubation period (7–10 days), the efficiency of vaccination at the population level ranged from 0.29 to 0.32. The duration of vaccine effectiveness is plotted in Fig.~\ref{fig5}(a). It is clear from the data presented in Fig.~\ref{fig5}(c) that different vaccination models will lead to different estimations of the DPC.
\begin{figure*}
\centering
\includegraphics[width=.8\textwidth]{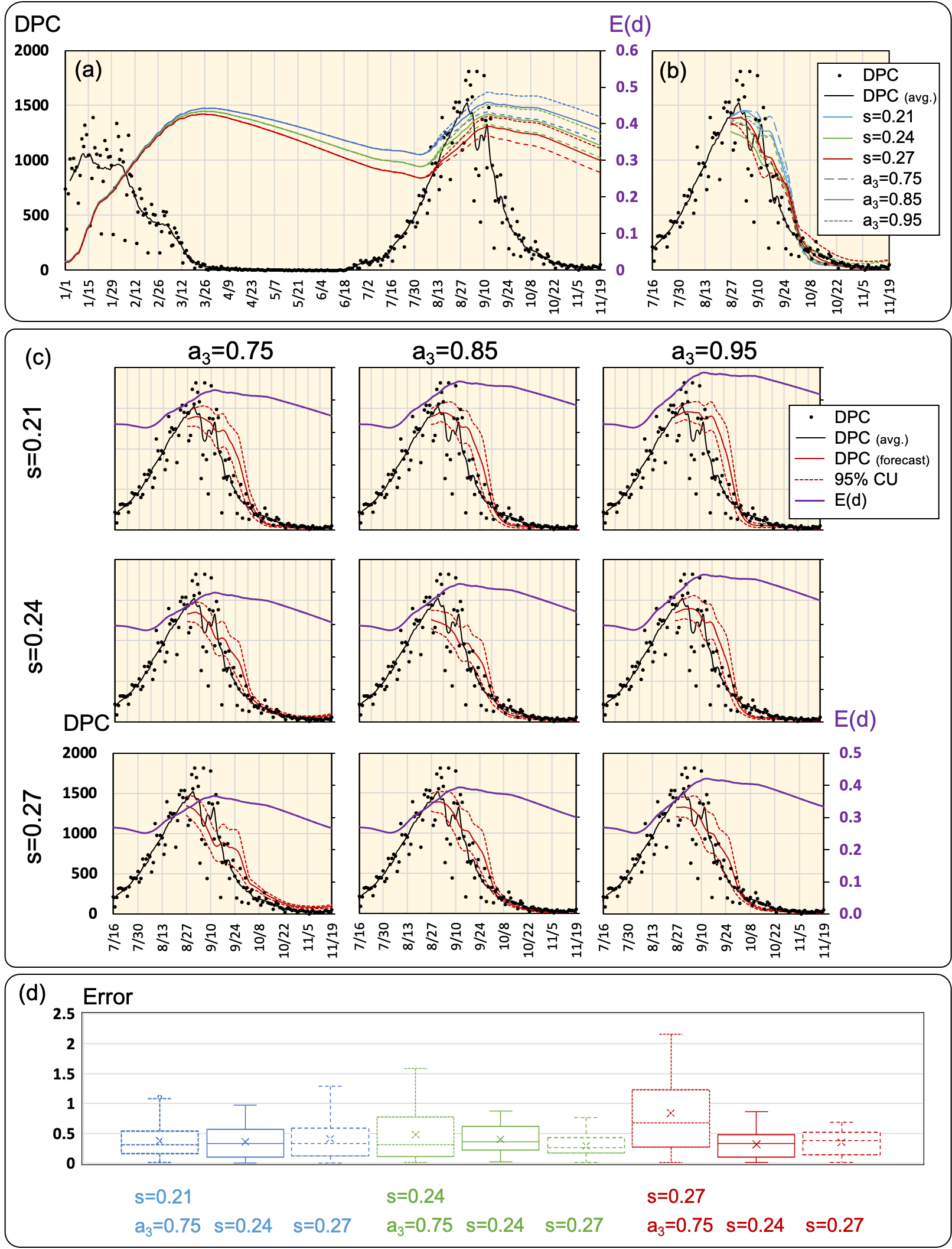}
\caption{(a) Vaccination effectiveness model (Eq.(\ref{eq1})) in Tel Aviv with different values of $s$ and $a_3$ along with DPC. (b) Forecasted DPC (7-day average) using different vaccination effectiveness models during the decay of the COVID-19 wave. (c) Detailed forecasted DPC data including the 95\% confidence interval and associated vaccination effectiveness model. (d) Error associated with the forecasts using different vaccination effectiveness models.}
\label{fig6}
\end{figure*}
\begin{figure*}
\centering
\includegraphics[width=0.9\textwidth]{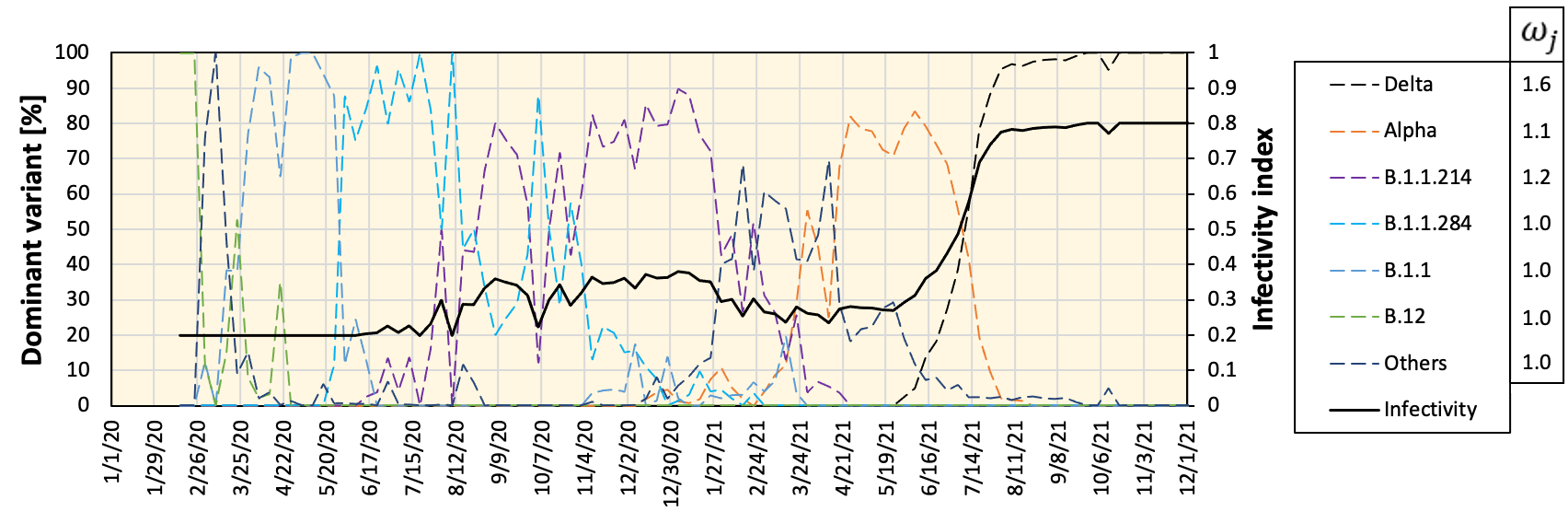}
\caption{Example of a variant infectivity index computed using data of viral variant measures in Tokyo with associated weight values representing relative infectivity.}
\label{fig7}
\end{figure*}

\begin{figure*}
\centering
\includegraphics[width=0.9\textwidth]{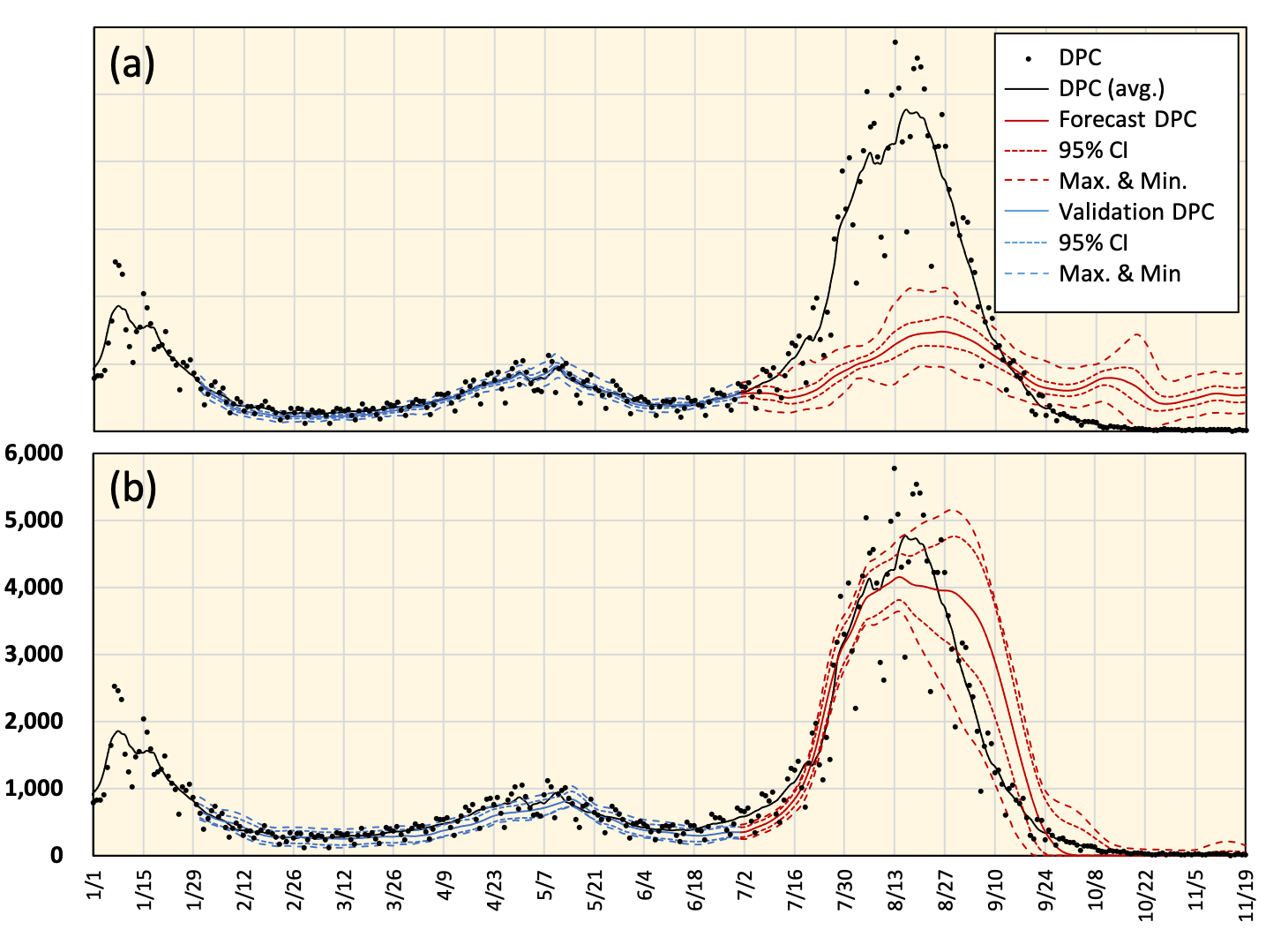}
\caption{Predicted DPC in Tokyo for the fifth wave with (a) all datasets (included in Table~\ref{tab1}) and (b) optimized datasets (only values of 1-1, 2-4, 5-1, 6-1, 7-1 from Table~\ref{tab1}) along with true values. Training data are from August 1, 2020 to July 30, 2021.}
\label{fig8}
\end{figure*}
\subsection{Input parameters for DPC}

After considering all input parameters, an extensive sensitivity study was conducted to minimize the input datasets. First, a single item was selected from each data set to minimize the total forecasting error. The seven selected inputs were then optimized to minimize possible number based on error. It was found that the optimized data inputs are those corresponded to the current DPC, mobility (transit station), model of the viral variant, twitter data \emph{nomikai}, and vaccination effectiveness. The variant infectivity computed for Tokyo is shown in Fig.~\ref{fig6}, with different weighting factors assigned to different viral variants. Figure~\ref{fig7} shows the estimated DPC for the given parameters in Tokyo. The first set was generated using all datasets whereas the second estimation was obtained using optimized data which matched with the most accurate observed values. As shown in Table~\ref{tab2}, the estimated DPC in Tokyo had error values of 0.23, 0.09, 0.78, and 0.36 for the spread, peak, decay, and all phases, respectively. The corresponding values for Osaka and Aichi were 0.24, 0.09, 0.41, 0.24, and 0.13, 0.16, 0.21, 0.17, respectively, which are highly consistent with the data of Tokyo. These results demonstrate that the viral variants model plays an important role during the spread phase in terms of starting time and peak value. In addition, the vaccination effectiveness model clearly contributes to the decay phase and can correctly forecast the rapid decay presented in the fifth COVID-19 wave in Japan. 

\begin{table*}
\centering
\footnotesize
\caption{Errors computed in the forecasts of separate phases of the fifth COVID-19 wave in Tokyo, Osaka, and Aichi with different data sets. Each experiment was conducted by excluding a single dataset (1-7) in Table~\ref{tab1}. \emph{None} demonstrates the case wherein all datasets are used, and \emph{Optimized} demonstrates the best scenario. Green and red colors are the lowest and highest error values, respectively.}
\setlength{\tabcolsep}{3pt}
\begin{tabular}{c|ccccccccc|c}
\hline
&& Ex.1 & Ex.2 & Ex.3 & Ex.4 & Ex.5 & Ex.6 & Ex.7 & None & Opt.$^*$\\
\hline\hline
\multirow{4}{*}{Tokyo} & Spread$^{**}$ & {\bf \color{purple}0.48}&0.30&0.26&{\bf \color{PineGreen}0.18}&0.41&0.53&0.43&0.44&0.23\\
\cline{2-11}
&Peak &0.64&0.61&0.45&0.56&{\bf \color{BrickRed}0.74}&0.65&0.63&0.68&{\bf\color{PineGreen}0.09}\\
\cline{2-11}
&Decay &0.79&1.06&{\bf \color{BrickRed}1.88}&0.63&{\bf\color{PineGreen}0.39}&0.61&0.78&0.53&0.78\\
\cline{2-11}
&Full wave &0.63&0.65&{\bf \color{BrickRed}0.85}&0.46&0.52&0.62&0.61&0.55&{\bf\color{PineGreen}0.36}\\
\hline\hline
\multirow{4}{*}{Osaka} & Spread &  {\bf\color{PineGreen}0.21} &0.39 &0.25 &0.26 &0.38 &{\bf \color{BrickRed}0.41} &0.35 &0.38 &0.24\\
\cline{2-11}
&Peak &0.81 &0.76 &0.69& 0.56 &{\bf \color{BrickRed}0.82} &0.75 &0.73& 0.75 &{\bf\color{PineGreen}0.09}\\
\cline{2-11}
&Decay & 0.55& 0.44 &0.42 &0.52 &0.36 &0.37 &{\bf\color{PineGreen}0.33}&{\bf \color{BrickRed}0.63} &0.41 \\
\cline{2-11}
&Full wave & 0.51 &0.53 &0.45 &0.45 &0.52 &0.53 &0.47 &{\bf \color{BrickRed}0.58} &{\bf\color{PineGreen}0.24} \\
\hline\hline
\multirow{4}{*}{Aichi} & Spread &   {\bf \color{BrickRed}0.83} &0.15 &0.71 &0.20& 0.71& 0.48 &0.14 &0.17& {\bf \color{PineGreen}0.13}\\
\cline{2-11}
&Peak &{\bf \color{BrickRed}0.86} &0.75 &0.55& 0.79& 0.55& 0.44& 0.49& 0.49&  {\bf \color{PineGreen}0.16}\\
\cline{2-11}
&Decay &  0.95 &0.85 &0.96 &0.65 &0.85& 0.92 &{\bf \color{BrickRed}1.08}& 0.60 & {\bf \color{PineGreen}0.21}  \\
\cline{2-11}
&Full wave &  {\bf \color{BrickRed}0.88}& 0.58& 0.74& 0.54& 0.70& 0.60& 0.57 &0.42&  {\bf \color{PineGreen}0.17} \\
\hline
\multicolumn{11}{l}{\it $^*$Opt.: optimized inputs are: 1-1, 2-4, 5-1,6-1 and 7-1 (see, Table~\ref{tab1}).} \\
\multicolumn{11}{l}{\it $^**$Spread (July), peak (Aug.), decay (Sept.), and all wave (July-Sept.).} \\
\end{tabular}
\label{tab2}
\end{table*}

\begin{figure}
\centering
\includegraphics[width=0.5\textwidth]{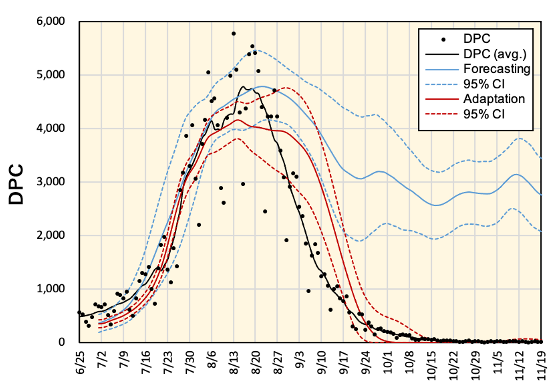}
\caption{Comparison of forecasting and adaptation models (shown in Fig. ~\ref{fig1}) for DPC in Tokyo. This demonstrates the effect of vaccination, which shows a weak spread phase and prolonged decay phase of the fifth COVID-19 wave.}
\label{fig9}
\end{figure}

\subsection{Adaptation of vaccination effectiveness modeling}

The estimation of new DPC using the combination of forecasting and adaptation models is shown in Fig.~\ref{fig8}. With the forecasting model alone, the spreading phase was highly consistent with real data; however, the decay phase was not. On the other hand, the combination of the two models resulted in more accurate results, especially in the decay phase, due to the application of the vaccination effectiveness acquired from the Tel Aviv data. The difference between the two models in the spread phase was marginal; however, it was significant in the decay phase. This tendency is similar to that of our previous study wherein the adaptation of new viral variants was discussed~\cite{ijerph18157799}. We found that different combinations of data as well as different time frames would lead to significant changes in the results, especially those for the long-term forecasting. For example, in the early stages of the pandemic and prior to the emergence of viral variants, the meteorological data was suggested to highly correlate with the incidence of infection~\cite{majumder2021systematic}. However, when new factors, such as vaccination effectiveness and viral variant infectivity, were considered, the importance of meteorological data was lessened. Table~\ref{tab2} demonstrates a brief assessment wherein a single dataset is excluded at a given time. This assessment was conducted in Tokyo, Osaka, and Aichi, and the learning period was from April 1, 2020 to June 30, 2021. Three time periods were included to demonstrate the spread (cases are increasing), peak (cases are at high values), and decay (cases are decreasing) of cases. Figure~\ref{fig9} demonstrates forecasting during a pandemic state wherein positives, serious cases, and deaths, for Tokyo, Osaka, and Aichi are forecasted. These results indicate that DPC and serious cases can be estimated with high accuracy while deaths are not. This might be attributable to the large variability of time-series data, which make it difficult for the LSTM network to learn the data pattern.

\begin{figure*}
\centering
\includegraphics[width=0.9\textwidth]{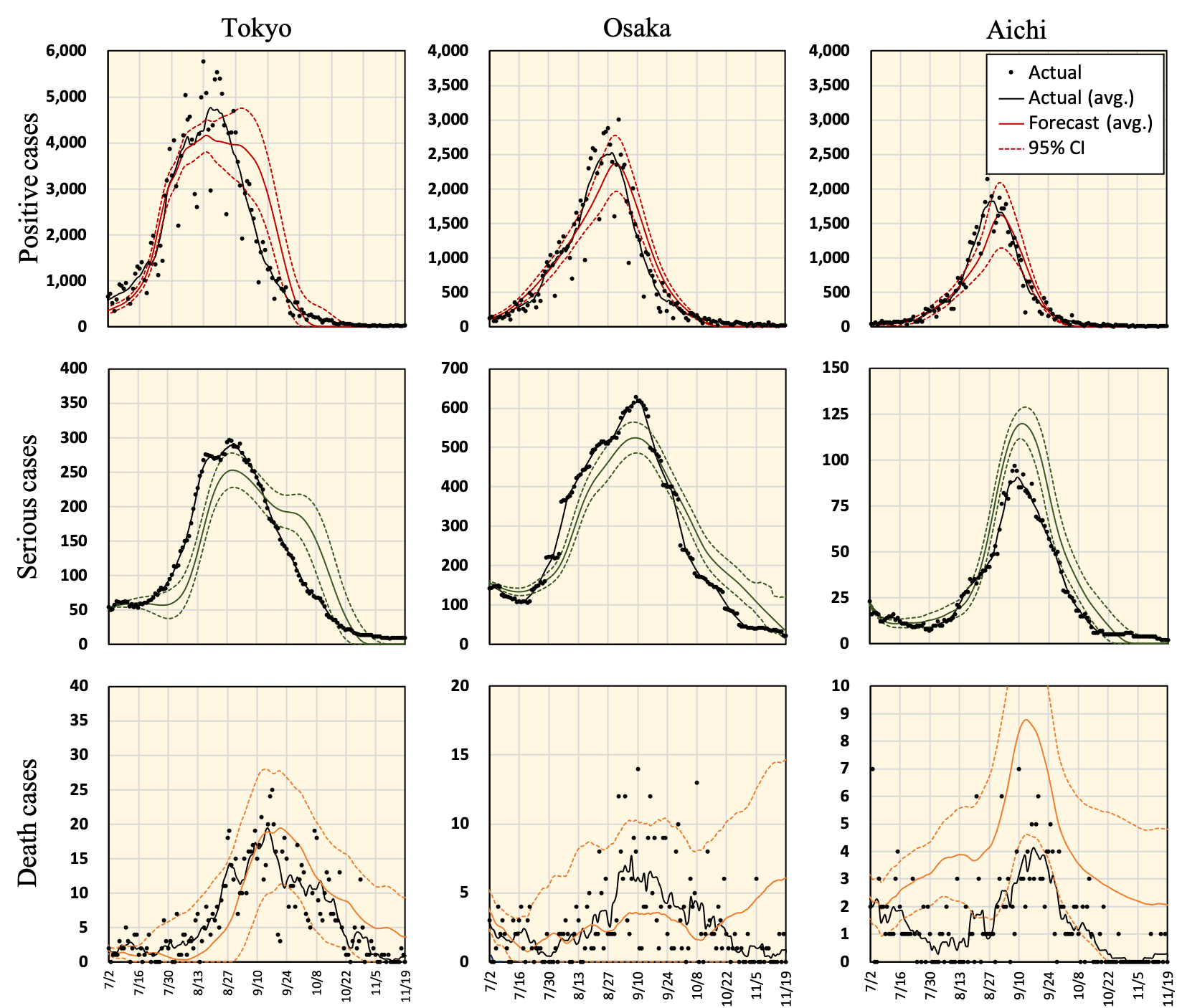}
\caption{Forecasting of DPC, serious cases, and deaths in Tokyo, Osaka, and Aichi with considering optimized input data and both forecasting and adaptation models.}
\label{fig10}
\end{figure*}

\section{Discussion}

In the early months of the COVID-19 pandemic, it was possible to estimate the DPC using only a small number of factors; however, after considering large-scale vaccination campaigns and the emergence of different variants, DPC estimation has become complicated. Regarding vaccination effectiveness, the effect of vaccination is not direct; hence, it needs to be carefully modeled by considering the variations in the vaccine type and potential waning of protection. Here, we present a method where the vaccination effect in one country can be projected to another country. Specifically, we have used vaccination data from Israel to train adaptation model (Fig.~\ref{fig1}(a)), which is used to adjust forecasting results of Japan obtained from forecasting model (Fig.~\ref{fig1}(b)). For new viral variants, it is crucial to model its infectivity to correctly estimate such that the trigger of new wave and potential peak can be correctly estimated.
Several deep learning approaches have been developed for COVID-19 forecasting using different factors such as current pandemic state, meteorological data, public mobility, and others. However, the trend of forecasting is changing with the wide administration of the vaccines and the potential higher infectivity of new variants. These new factors are hardly being used in previous models due to the data limitations. We studied forecasting using different set of inputs that demonstrate varied factors discussed in the literature such as public mobility and behavior, meteorological data, vaccination effectiveness, and viral variant infectivity. The results demonstrate that different parameters have different implications along a given time course. Therefore, the training data should be carefully selected to obtain highly accurate long-term forecasts. We presented a feasible method to project the vaccination effectiveness obtained from another country and a model to manage the change in the infectivity of viral variants. As many countries have vaccination rates still below the target threshold for herd immunity set by the World Health Organization, it would be useful to validate the potential risks and forecast future waves of infection using data from other regions with high vaccination rates. 

While the present study demonstrate a method to model vaccination effectiveness and viral variants for accurate forecasting of DPC, it has several limitations to be listed. The data used here are obtained from two countries where mRNA-based vaccines are used. The performance of the proposed model is unknown if the the vaccine development technology is different. Also, the percentage of viral variant used in this study is based on relatively small number of samples that might not be a representative of real distribution.

\section{Conclusion}

In this study, we demonstrate a new framework for including infectivity variation caused by different viral variants and potential protection caused by vaccination effectiveness. These factors in addition to other known correlated data such as meteorological data, public mobility and others are combined in two successive LSTM models. The first model is a local scope model (forecasting) that is trained by local data measurements. The second model is a global scope model (adaptation) that can be trained using external data and used to adjust forecasting results. This approach demonstrate high accuracy results to forecast DPC in three regions of Japan with vaccination data obtained from Israel. This approach can be used to forecast DPC in countries with low vaccination rate using measurements at other countries with high vaccination rate. Therefore, the scope of potential usage is large as average global vaccination rate is still low in most countries.

\section*{Acknowledgment}

The authors would like to thank Professor Masashi Toyoda (Institute of Industrial Science, The University of Tokyo, Japan) for generating processed Twitter data based on raw data provided by NTT Data, INC (Japan).

\bibliography{Refs1}

\end{document}